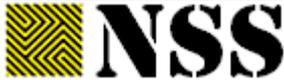

# The Neutrosophic Entropy and its Five Components

**Vasile Pătraşcu**

Tarom Information Technology, 224F Bucurestilor Road, Otopeni, 075150, Romania. E-mail: patrascu.v@gmail.com

**Abstract.** This paper presents two variants of penta-valued representation for neutrosophic entropy. The first is an extension of Kaufmann's formula and the second is an extension of Kosko's formula.

Based on the primary three-valued information represented by the degree of truth, degree of falsity and degree of neutrality there are built some penta-valued representations that better highlights some specific features of neutrosophic entropy. Thus, we highlight five features of neutrosophic uncertainty such as ambiguity, ignorance, contradiction, neutrality and saturation. These five features are supplemented until a seven partition of unity by adding two features of neutrosophic certainty such as truth and falsity.

The paper also presents the particular forms of neutrosophic entropy obtained in the case of bifuzzy representations, intuitionistic fuzzy representations, paraconsistent fuzzy representations and finally the case of fuzzy representations.

## 1 Introduction

Neutrosophic representation of information was proposed by Smarandache [10], [11], [12] as an extension of fuzzy representation proposed by Zadeh [16] and intuitionistic fuzzy representation proposed by Atanassov [1], [2]. Primary neutrosophic information is defined by three parameters: degree of truth $\mu$, degree of falsity $\nu$ and degree of neutrality $\omega$.

Fuzzy representation is described by a single parameter, degree of truth $\mu$, while the degree of falsity $\nu$ has a default value calculated by negation formula:

$$\nu = 1 - \mu \qquad (1.1)$$

and the degree of neutrality has a default value that is $\omega = 0$.

Fuzzy intuitionistic representation is described by two explicit parameters, degree of truth $\mu$ and degree of falsity $\nu$, while the degree of neutrality has a default value that is $\omega = 0$.

Atanassov considered the incomplete variant taking into account that $\mu + \nu \leq 1$. This allowed defining the index of ignorance (incompleteness) with the formula:

$$\pi = 1 - \mu - \nu \qquad (1.2)$$

thus obtaining a consistent representation of information, because the sum of the three parameters is 1, namely:

$$\mu + \pi + \nu = 1 \qquad (1.3)$$

Hence, we get for neutrality the value $\omega = 0$.

For paraconsistent fuzzy information where $\mu + \nu \geq 1$, the index of contradiction can be defined:

$$\kappa = \mu + \nu - 1 \qquad (1.4)$$

and for neutrality it results: $\omega = 0$.

For bifuzzy information that is defined by the pair $(\mu, \nu)$, the net truth, the index of ignorance (incompleteness), index of contradiction and index of ambiguity can be defined, by:

$$\tau = \mu - \nu \qquad (1.5)$$

$$\pi = 1 - \min(\mu + \nu, 1) \qquad (1.6)$$

$$\kappa = \max(\mu + \nu, 1) - 1 \qquad (1.7)$$

$$\alpha = 1 - |\mu - \nu| - |\mu + \nu - 1| \qquad (1.8)$$

Among of these information parameters the most important is the construction of some measures for information entropy or information uncertainty. This paper is dedicated to the construction of neutrosophic entropy formulae.

In the next, section 2 presents the construction of two variants of the neutrosophic entropy. This construction is based on two similarity formulae; Section 3 presents a penta-valued representation of neutrosophic entropy based on ambiguity, ignorance, contradiction, neutrality and saturation; Section 4 outlines some conclusions.

## 2 The Neutrosophic Entropy

For neutrosophic entropy, we will trace the Kosko idea for fuzziness calculation [5]. Kosko proposed to measure this information feature by a similarity function between the distance to the nearest crisp element and the distance to the farthest crisp element. For neutrosophic information the two crisp elements are $(1,0,0)$ and $(0,0,1)$. We consider the following vector: $V = (\mu - \nu, \mu + \nu - 1, \omega)$. For $(1,0,0)$ and $(0,0,1)$ it results: $V_T = (1,0,0)$ and $V_F = (-1,0,0)$. We will compute the distances:

$$D(V, V_T) = |\mu - \nu - 1| + |\mu + \nu - 1| + \omega \quad (2.1)$$

$$D(V, V_F) = |\mu - \nu + 1| + |\mu + \nu - 1| + \omega \quad (2.2)$$

The neutrosophic entropy will be defined by the similarity between these two distances.

Using the Czekanowskyi formula [3] it results the similarity $S_C$ and the neutrosophic entropy $E_C$:

$$S_C = 1 - \frac{|D(V, V_T) - D(V, V_F)|}{D(V, V_T) + D(V, V_F)} \quad (2.3)$$

$$E_C = 1 - \frac{|\mu - \nu|}{1 + |\mu + \nu - 1| + \omega} \quad (2.4)$$

or in terms of $\tau, \omega, \pi, \kappa$:

$$E_C = 1 - \frac{|\tau|}{1 + \pi + \kappa + \omega} \quad (2.5)$$

The neutrosophic entropy defined by (2.4) can be particularized for the following cases:

For $\omega = 0$, it result the bifuzzy entropy, namely:

$$E_C = 1 - \frac{|\mu - \nu|}{1 + |\mu + \nu - 1|} \quad (2.6)$$

For $\omega = 0$ and $\mu + \nu \leq 1$ it results the intuitionistic fuzzy entropy proposed by Patrascu [8], namely:

$$E_C = 1 - \frac{|\mu - \nu|}{1 + \pi} \quad (2.7)$$

For $\omega = 0$ and $\mu + \nu \geq 1$ it results the paraconsistent fuzzy entropy, namely:

$$E_C = 1 - \frac{|\mu - \nu|}{1 + \kappa} \quad (2.8)$$

For $\mu + \nu = 1$ and $\omega = 0$, it results the fuzzy entropy proposed by Kaufmann [4], namely:

$$E_C = 1 - |2\mu - 1| \quad (2.9)$$

Using the Ruzicka formula [3] it result the formulae for the similarity $S_R$ and the neutrosophic entropy $E_R$:

$$S_R = 1 - \frac{|D(V, V_T) - D(V, V_F)|}{\max(D(V, V_T), D(V, V_F))} \quad (2.10)$$

$$E_R = 1 - \frac{2|\mu - \nu|}{1 + |\mu - \nu| + |\mu + \nu - 1| + \omega} \quad (2.11)$$

or its equivalent form:

$$E_R = \frac{1 - |\mu - \nu| + |\mu + \nu - 1| + \omega}{1 + |\mu - \nu| + |\mu + \nu - 1| + \omega} \quad (2.12)$$

or in terms of $\tau, \omega, \pi, \kappa$:

$$E_R = 1 - \frac{2|\tau|}{1 + |\tau| + \pi + \kappa + \omega} \quad (2.13)$$

The neutrosophic entropy defined by (2.12) can be particularized for the following cases:

For $\omega = 0$, it results the bifuzzy entropy proposed by Patrascu [7], namely:

$$E_R = \frac{1 - |\mu - \nu| + |\mu + \nu - 1|}{1 + |\mu - \nu| + |\mu + \nu - 1|} \quad (2.14)$$

For $\omega = 0$ and $\mu + \nu \leq 1$ it results the intuitionistic fuzzy entropy proposed by Szmidt and Kacprzyk [14], [15], explicitly:

$$E_R = \frac{1 - |\mu - \nu| + \pi}{1 + |\mu - \nu| + \pi} \quad (2.15)$$

For $\omega = 0$ and $\mu + \nu \geq 1$ it results the paraconsistent fuzzy entropy, explicitly:

$$E_R = \frac{1 - |\mu - \nu| + \kappa}{1 + |\mu - \nu| + \kappa} \quad (2.16)$$

For $\mu + \nu = 1$ and $\omega = 0$, it results the fuzzy entropy proposed by Kosko [5], namely:

$$E_R = \frac{1 - |2\mu - 1|}{1 + |2\mu - 1|} \quad (2.17)$$

We notice that the neutrosophic entropy is a strictly decreasing function in $|\mu - \nu|$ and non-decreasing in $\omega$ and in $|\mu + \nu - 1|$.

The neutrosophic entropy verify the following conditions:

(i) $E(\mu,\omega,\nu) = 0$ if $(\mu,\omega,\nu)$ is a crisp value, namely if $(\mu,\omega,\nu) \in \{(1,0,0),(0,0,1)\}$.

(ii) $E(\mu,\omega,\nu) = 1$         if $\mu = \nu$.

(iii) $E(\mu,\omega,\nu) = E(\nu,\omega,\mu)$

(iv) $E(\mu,\omega_1,\nu) \leq E(\mu,\omega_2,\nu)$     if $\omega_1 \leq \omega_2$.

(v) $E(\tau_1,\omega,\pi,\kappa) < E(\tau_2,\omega,\pi,\kappa)$   if $|\tau_1| > |\tau_2|$.

(vi) $E(\tau,\omega_1,\pi,\kappa) \leq E(\tau,\omega_2,\pi,\kappa)$   if $\omega_1 \leq \omega_2$.

(vii) $E(\tau,\omega,\pi_1,\kappa) \leq E(\tau,\omega,\pi_2,\kappa)$   if $\pi_1 \leq \pi_2$.

(viii) $E(\tau,\omega,\pi,\kappa_1) \leq E(\tau,\omega,\pi,\kappa_2)$ if $\kappa_1 \leq \kappa_2$.

## 3 Penta-valued Representation of Neutrosophic Entropy

The five components of neutrosophic entropy will be: ambiguity $a$, ignorance $u$, contradiction $c$, neutrality $n$ and saturation $s$ [6]. We construct formulas for these features both for the variant defined by formula (2.4) and for the variant defined by formula (2.12). For each decomposition, among the four components $u,n,s,c$, always two of them will be zero.

In the neutrosophic cube, we consider the entropic rectangle defined by the points: $U = (0,0,0)$, $N = (0,1,0)$, $S = (1,1,1)$, $C = (1,0,1)$.

In addition, we consider the point $A = (0.5,0,0.5)$ which is located midway between the point $U$ (*unknown*) and the point $C$ (*contradiction*).

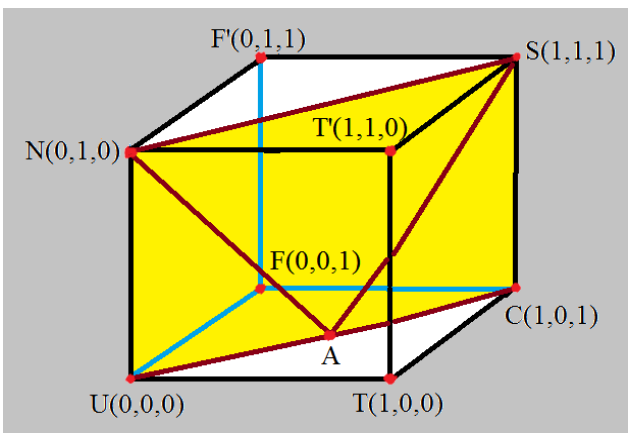

Figure 1. The neutrosophic cube TUFCT'NF'S and its entropic rectangle UNSC.

Also, this point is located midway between the point $F$ (*false*) and the point $T$ (*true*). In other words, the point $A$ (*ambiguous*) represents the center of the Belnap square TUFC (*true-uknown-false-contradictory*).

If the projection of the point $(\mu,\omega,\nu)$ on the rectangle UNSC (*unknown-neutral-saturated-contradictory*) is inside the triangle UNA (*unknown-neutral-ambiguous*) then saturation and contradiction will be zero $(s = c = 0)$, if the projection is inside the triangle ANS (*ambigouous-neutral-saturated*) then ignorance and contradiction will be zero $(u = c = 0)$ and if the projection is inside the triangle ASC (*ambiguous-saturated-contradictory*) then ignorance and neutrality will be zero $(u = n = 0)$.

We consider first the first version defined by formula (2.4), namely:

$$E_C = 1 - \frac{|\mu - \nu|}{1 + |\mu + \nu - 1| + \omega} \quad (3.1)$$

The five features must verify the following condition:

$$a + u + c + n + s = E_C \quad (3.2)$$

First, we start with ambiguity formula defined by:

$$a = \frac{1 - |\mu - \nu| - |\mu + \nu - 1|}{1 + |\mu + \nu - 1| + \omega} \quad (3.3)$$

The prototype for the ambiguity is the point $(0.5,0,0.5)$ and the formula (3.3) defines the similarity between the points $(\mu,\omega,\nu)$ and $(0.5,0,0.5)$.

Then, the other four features will verify the condition:

$$u + c + n + s = \frac{2|\mu + \nu - 1| + \omega}{1 + |\mu + \nu - 1| + \omega} \quad (3.4)$$

We analyze three cases depending on the order relation among the three parameters $\omega,\pi,\kappa$ where $\pi$ is the bifuzzy *ignorance* and $\kappa$ is the bifuzzy *contradiction* [9]. It is obvious that:

$$\pi \cdot \kappa = 0 \quad (3.5)$$

**Case (I)**

$$\pi \geq \omega \geq \kappa = 0 \quad (3.6)$$

It results for ignorance and contradiction these formulas:

$$u = \frac{2\pi - 2\omega}{1 + \omega + \pi + \kappa} \quad (3.7)$$

$$c = 0 \quad (3.8)$$

and for saturation and neutrality:

$$s = 0 \quad (3.9)$$

$$n = \frac{3\omega}{1+\omega+\pi+\kappa} \quad (3.10)$$

**Case (II)**

$$\kappa \geq \omega \geq \pi = 0 \quad (3.11)$$

It results for ignorance and contradiction these formulas:

$$u = 0 \quad (3.12)$$

$$c = \frac{2\kappa - 2\omega}{1+\omega+\pi+\kappa} \quad (3.13)$$

then for neutrality and saturation it results:

$$n = 0 \quad (3.14)$$

$$s = \frac{3\omega}{1+\omega+\pi+\kappa} \quad (3.15)$$

**Case (III)**

$$\omega \geq \max(\pi, \kappa) \quad (3.16)$$

It results for ignorance and contradiction these values:

$$u = 0 \quad (3.17)$$

$$c = 0 \quad (3.18)$$

Next we obtain:

$$s + n = \frac{(\omega - \pi - \kappa) + 3\pi + 3\kappa}{1+\omega+\pi+\kappa} \quad (3.19)$$

The sum (3.19) can be split in the following manner:

$$n = \frac{\frac{\omega-\pi-\kappa}{2} + 3\pi}{1+\omega+\pi+\kappa} \quad (3.20)$$

$$s = \frac{\frac{\omega-\pi-\kappa}{2} + 3\kappa}{1+\omega+\pi+\kappa} \quad (3.21)$$

Combining formulas previously obtained, it results the final formulas for the five components of the neutrosophic entropy defined by (3.1):

*ambiguity*

$$a = \frac{1 - |\mu - \nu| - \pi - \kappa}{1+\omega+\pi+\kappa} \quad (3.22)$$

*ignorance*

$$u = 2 \cdot \frac{\max(\pi,\omega) - \omega}{1+\omega+\pi+\kappa} \quad (3.23)$$

The prototype for *ignorance* is the point $(0,0,0)$ and formula (3.23) defines the similarity between the points $(\mu,\omega,\nu)$ and $(0,0,0)$.

*contradiction*

$$c = 2 \cdot \frac{\max(\kappa,\omega) - \omega}{1+\omega+\pi+\kappa} \quad (3.24)$$

The prototype for *contradiction* is the point $(1,0,1)$ and formula (3.24) defines the similarity between the points $(\mu,\omega,\nu)$ and $(1,0,1)$.

*neutrality*

$$n = \frac{\frac{\max(\omega-\pi-\kappa,0)}{2} + 3\min(\omega,\pi)}{1+\omega+\pi+\kappa} \quad (3.25)$$

The prototype for *neutraliy* is the point $(0,1,0)$ and formula (3.25) defines the similarity between the points $(\mu,\omega,\nu)$ and $(0,1,0)$.

*saturation*

$$s = \frac{\frac{\max(\omega-\pi-\kappa,0)}{2} + 3\min(\omega,\kappa)}{1+\omega+\pi+\kappa} \quad (3.26)$$

The prototype for *saturation* is the point $(1,1,1)$ and formula (3.26) defines the similarity between the points $(\mu,\omega,\nu)$ and $(1,1,1)$.

In addition we also define:

*The index of truth*

$$t = \frac{\max(\mu,\nu) - \nu}{1+\omega+\pi+\kappa} \quad (3.27)$$

The prototype for the *truth* is the point $(1,0,0)$ and formula (3.27) defines the similarity between the points $(\mu,\omega,\nu)$ and $(1,0,0)$.

*The index of falsity*

$$f = \frac{\max(\nu,\mu) - \mu}{1+\omega+\pi+\kappa} \quad (3.28)$$

The prototype for the *falsity* is the point $(0,0,1)$ and formula (3.28) defines the similarity between the points $(\mu,\omega,\nu)$ and $(0,0,1)$.

Thus, we get the following hepta-valued partition for the neutrosophic information:

$$t + f + a + u + c + n + s = 1 \quad (3.29)$$

Formula (3.29) shows that neutrosophic information can be structured so that it is related to a logic where the information could be: true, false, ambiguous, unknown, contradictory, neutral or saturated [6] (see Figure 2).

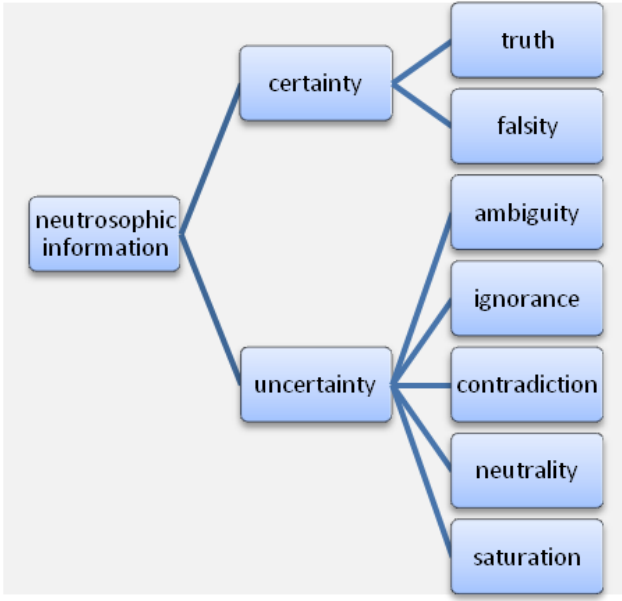

Figure 2. The structure of the neutrosophic information.

Next, we also deduce the five components for the variant defined by the formula (2.12).

Using (1.5) formula (2.12) becomes:

$$E_R = 1 - \frac{2|\tau|}{1+|\tau|+\pi+\kappa+\omega} \tag{3.30}$$

or

$$E_R = \frac{1-|\tau|+\pi+\kappa+\omega}{1+|\tau|+\pi+\kappa+\omega} \tag{3.31}$$

In this case, the five formulas are obtained from formulas (3.22) - (3.26) by changing the denominator, thus:

*ambiguity*

$$a = \frac{1-|\tau|-\pi-\kappa}{1+|\tau|+\omega+\pi+\kappa} \tag{3.32}$$

*ignorance*

$$u = \frac{2\max(\pi-\omega,0)}{1+|\tau|+\omega+\pi+\kappa} \tag{3.33}$$

*contradiction*

$$c = \frac{2\max(\kappa-\omega,0)}{1+|\tau|+\omega+\pi+\kappa} \tag{3.34}$$

*neutrality*

$$n = \frac{\frac{\max(\omega-\pi-\kappa,0)}{2} + 3\min(\omega,\pi)}{1+|\tau|+\omega+\pi+\kappa} \tag{3.35}$$

*saturation*

$$s = \frac{\frac{\max(\omega-\pi-\kappa,0)}{2} + 3\min(\omega,\kappa)}{1+|\tau|+\omega+\pi+\kappa} \tag{3.36}$$

From (3.27) and (3.28) it results:

*Index of truth*

$$t = \frac{2\max(\tau,0)}{1+|\tau|+\omega+\pi+\kappa} \tag{3.37}$$

*Index of falsity*

$$f = \frac{2\max(-\tau,0)}{1+|\tau|+\omega+\pi+\kappa} \tag{3.38}$$

Also in this case, the 7 parameters verify the condition of partition of unity defined by formula (3.29).

For bifuzzy information when $\omega=0$, neutrality and saturation are zero and formula (3.29) becomes:

$$t+f+a+u+c = 1 \tag{3.39}$$

Thus, we obtained for bifuzzy information a penta-valued representation. We can conclude that the bifuzzy information is related to a penta-valued logic where the information could be: true, false, ambiguous, unknown and contradictory [6] (see Figure 3).

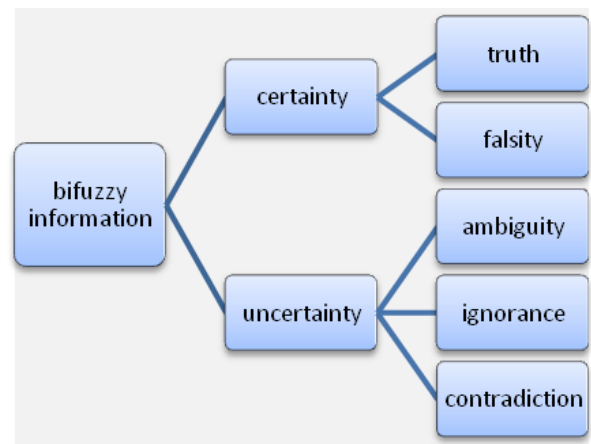

Figure 3. The structure of the bifuzzy information.

For intuitionistic fuzzy information when $\omega = 0$ and $\mu + \nu \leq 1$, neutrality, saturation and contradiction are zero and formula (3.29) becomes:

$$t + f + a + u = 1 \qquad (3.40)$$

Thus, we obtained for intuitionistic fuzzy information a tetra-valued representation. We can conclude that the intuitionistic fuzzy information is related to a tetra-valued logic where the information could be: true, false, ambiguous and unknown [6] (see Figure 4).

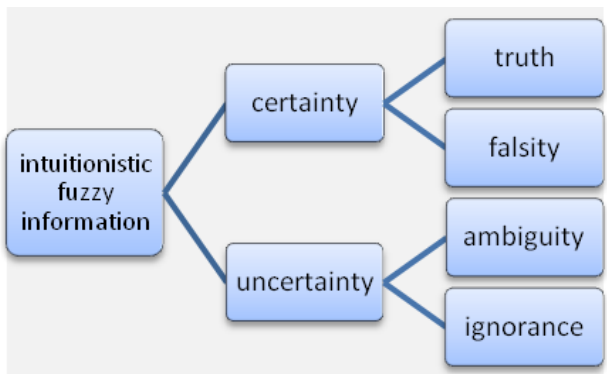

Figure 4. The structure of the intuitionistic fuzzy information.

For paraconsistent fuzzy information when $\omega = 0$ and $\mu + \nu > 1$, neutrality, saturation and ignorance are zero and formula (3.29) becomes:

$$t + f + a + c = 1 \qquad (3.41)$$

The paraconsistent fuzzy information is related to a tetra-valued logic where the information could be: true, false, ambiguous and contradictory [6] (see Figure 5).

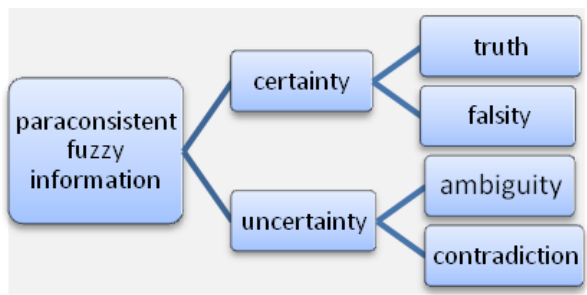

Figure 5. The structure of the paraconsistent fuzzy information.

For fuzzy information when $\omega = 0$ and $\mu + \nu = 1$, neutrality, saturation ignorance and contradiction are zero and formula (3.29) becomes:

$$t + f + a = 1 \qquad (3.42)$$

Thus, we obtained for fuzzy information a three-valued representation. We can conclude that the fuzzy information is related to a three-valued logic where the information could be: true, false and ambiguous [6] (see Figure 6).

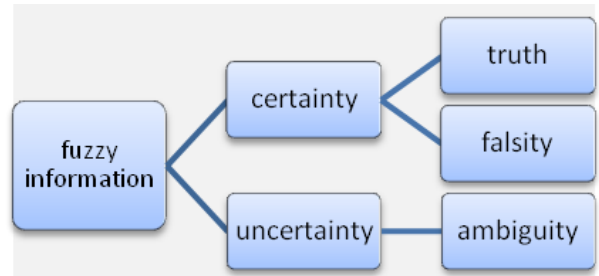

Figure 6. The structure of the fuzzy information.

## 4 Conclusion

In this article, there are constructed two formulas for neutrosophic entropy calculating. For each of these, five components are defined and they are related to the following features of neutrosophic uncertainty: ambiguity, ignorance, contradiction, neutrality and saturation. Also, two components are built for neutrosophic certainty: truth and falsity. All seven components, five of the uncertainty and two of certainty form a partition of unity. Building these seven components, primary neutrosophic information is transformed in a more nuanced representation.